\documentclass{IOS-Book-Article}

\usepackage{mathptmx}
\usepackage{soul}\setuldepth{article}
 
\usepackage{graphicx}
\usepackage{xcolor}
\usepackage[pagebackref,hidelinks,colorlinks = true,
            colorlinks = true,
            linkcolor = blue,
            urlcolor  = blue,
            citecolor = blue,
            anchorcolor = blue]{hyperref}

\makeatletter
\renewcommand{\@biblabel}[1]{(#1)}
\makeatother
\usepackage{url}
\usepackage{ragged2e}
\def\hb{\hbox to 10.7 cm{}}

\begin{document}

\pagestyle{plain}

\newcommand{\MYhref}[3][blue]{\href{#2}{\color{#1}{#3}}}
\begin{frontmatter}              % The preamble begins here.

%\pretitle{Pretitle}
\title{Ontology and Cognitive Outcomes}

%\markboth{}{Draft Submitted to FIOS 2020\hb}
%\subtitle{Subtitle}

\author[A]{\fnms{David} \snm{Limbaugh}%
\thanks{Corresponding author, Department of Philosophy, University at Buffalo, 135 Park Hall, Buffalo, NY 14260; E-mail: dglimbau@buffalo.edu}},
\author[B]{\fnms{Jobst} \snm{Landgrebe}},
\author[A]{\fnms{David} \snm{Kasmier}},
\author[C]{\fnms{Ronald} \snm{Rudnicki}},
\author[D]{\fnms{James} \snm{Llinas}},
and
\author[A]{\fnms{Barry} \snm{Smith}}

%\runningauthor{B.P. Manager et al.}
\address[A]{Department of Philosophy, University at Buffalo}
\address[B]{Cognotekt GmbH, Cologne, Germany}
\address[C]{CUBRC, Inc.}
\address[D]{Department of Industrial and Systems Engineering, University at Buffalo}

\begin{abstract}
Here we understand `intelligence' as referring to items of knowledge collected for the sake of assessing and maintaining national security. The intelligence community (IC) of the United States (US) is a community of organizations that collaborate in collecting and processing intelligence for the US. The IC relies on human-machine-based analytic strategies that 1) access and integrate vast amounts of information from disparate sources, 2) continuously process this information, so that, 3) a maximally comprehensive understanding of world actors and their behaviors can be developed and updated. Herein we describe an approach to utilizing outcomes-based learning (OBL) to support these efforts that is based on an ontology of the cognitive processes performed by intelligence analysts. Of particular importance to this Cognitive Process Ontology CPO is the class \textit{Representation that is Warranted}. A representation of this sort is descriptive in nature and deserving of trust in its veridicality. The latter holds because a Representation that is Warranted is always produced by a process that was vetted (or successfully designed) to reliably produce veridical representations. As such, Representations that are Warranted are what in other contexts we might refer to as `items of knowledge'. 
\end{abstract}

\end{frontmatter}
%\markboth{May 202\hb}{May 202\hb}
%\thispagestyle{empty}
%\pagestyle{empty}

%todo Replace names with citations
%todo Check all in text citations
%todo fix references: alphabitize and so on.

Type: original research article

\pagebreak

\section{Introduction}
We define an `ontology' as a controlled vocabulary of terms that represent real entities, where the terms in the ontology are used to semantically enhance bodies of data in such a way as to make even highly heterogeneous data more consistently accessible to computers (Salmen et al. 2011). Herein we discuss the Cognitive Process Ontology, where by `cognitive process' or `act of cognition' we mean a process that creates, modifies, or has as participant some item of knowledge. We describe the role of such an ontology and describe the role it can play in a system of outcomes-based assessments of analytical workflows, and show how this system may then be applied to improve collaboration within the intelligence community. By `analytical workflow' we understand a series of steps performed by humans or teams of humans using computers to transform heterogeneous data and information into decisions.

The data in question can be in any domain, but we focus here on the \textit{intelligence} domain, where workflows of the relevant sort have been intensively studied. The sense of `intelligence' used for our purposes here is that of an item of knowledge that is of strategic importance to the success of some enterprise, as in `business intelligence' or `military intelligence'. More specifically, we mean by `intelligence' those items of knowledge that relate to the national security of the United States (US), and which are sought after by the US Intelligence Community (IC). Hence, when we refer to `intelligence analyst' we refer to someone who -- as a vocation -- analyzes items of knowledge related to the national security of the US; \textit{a fortiori} `intelligence analysis' henceforth refers to the analyzing of items of knowledge as it relates to national security.

\section{Background}
As technology has advanced, the volume of data and information being made available to analysts has grown exponentially. Consequently, the Department of Defense (DoD) and the Intelligence Community (IC) must adapt to performing intelligence-related analysis on ever-growing amounts of distributed data (Lawrence 2012). By `intelligence-related analysis' we mean more precisely the processing, exploitation, and dissemination (PED) portions of the IC's \textit{intelligence cycle}. This cycle begins with an \textit{intelligence direction}, which is established on the basis of a need for information. This is followed by \textit{data gathering} from the operational environment. Gathered data is then \textit{processed} and \textit{exploited} for the sake of new information, and it is here that the analytic workflows that are the focus of this paper are located. Finally, new information is \textit{disseminated} to those who need it, which results in the cycle beginning again (Rosenbach and Peritz, 2012). This cycle is responsible for producing intelligence ``at all levels of national security -- from the war-fighter on the ground to the President in Washington'' (Rosenbach and Peritz, 2012).

As the increasing use of technology increases the amount of data flowing through the intelligence cycle, this creates a situation in which technology becomes crucial also to the processing of these data, but this comes with challenges as a result of the fact that human analysts must also be involved (Marcus 2018; Landgrebe and Smith 2019a). Algorithms are intended to assist humans in handling ever larger amounts of data. However, the effectiveness of algorithms -- even machine learning algorithms -- is limited by how quickly important results can enter the human decision chain. This, then, requires that the outputs of algorithms be discoverable by, and comprehensible to, humans. This in turn requires that data formats, and data coding and tagging systems, be used that make human access to and control over data analysis easier and the results of queries more readable. Furthermore, because an algorithm is merely a set of rules, it cannot create meaningful output data without meaningful input data. 

The human effort needed to address the above challenges, which are created by the shear amount of data being collected, has widened the gap between data collection and data exploitation. The Director of National Intelligence has in consequence asserted:
\begin{quote}
Closing the gap between decisions and data collection is a top priority for the Intelligence Community (IC). The pace at which data are generated and collected is increasing exponentially -- and the IC workforce available to analyze and interpret this all-source, cross-domain data is not. Leveraging artificial intelligence, automation, and augmentation technologies to amplify the effectiveness of our workforce will advance mission capability and enhance the IC's ability to provide needed data interpretation to decision makers (ODNI 2019).
\end{quote}
The upshot is that there is too much data and not enough human power to make effective use thereof.

\section{A Proposal}
The challenge is to close the gap between data collection and decision making. We propose to address this challenge by moving our attention one level higher, to the intelligence process itself. More specifically we propose to collect and to ontologically annotate data, created by humans or machines during the intelligence cycle in order to

\begin{enumerate}

\item enable better interoperability between different intelligence institutions and
\item to provide an improved foundation for the measurement of process effectiveness and thereby enable continuous process improvements.
\end{enumerate} 

\noindent Note that such annotated data cannot be used to automate portions of
the intelligence cycle hitherto performed by human beings because the mental
processes involved cannot be mathematically modeled (Landgrebe and Smith 2019b).

An ontological annotation of data means that primary or secondary process data are linked to ontological terms (manually or automatically). Primary data are process input or output data, inputs such as observation data or documents that are fed into intelligence process and outputs such as analysis results and recommendations that are generated by the intelligence process. Secondary data are data describing process activities and their relationships to each other. To perform such annotations, we have created the Cognitive Process Ontology (CPO), consisting of terms representing the cognitive processes -- kinds of mental processes -- used by analysts, such as `cognitive process of comparing', `cognitive process of inferring', `cognitive process of association', and `analysis of competing hypotheses' (ACH). Terms in CPO also represent the mental outputs of such processes, such as `Representation that is Believed' and `Representation that is Warranted'. 

Further ancillary terms are also included, representing what in the reality-outside-of-the-mind guides mental processes. These include terms, such as `indicator', which refers to some portion of reality that, if known about, changes one's estimation that some other portion of reality exists, has existed, or will exist. For example, knowing that a certain person visited a warehouse containing certain chemicals might be an indicator that an explosive device will soon exist.
 
Ancillary terms of this sort can be used to link CPO, which we can think of as an internally directed (directed at the mental) ontology, to other, externally directed ontologies such as those that comprise the Common Core Ontologies (CCO) ecosystem (CUBRC 2019a). For example, we can use CPO-CCO combinations to create compound terms (and corresponding complex graphs) such as `Representation that is Warranted about Planned Missile Launch' or `Evidence of Kinetic Kill Maneuver'. Aggregates of such terms can then be used to represent the components of investigative processes and process pipelines leading from data ingestion to informed decisions, including the outputs of such pipelines, for example in the form of predictions of real-world events (Chapman et al. 2020).

%Once we can represent investigative processes, including a granular representation of component mental processes, we are then able to tag such outputs of processes in light of how they contribute to decision making and fulfilling mission objectives. For instance, to what extent performing one cognitive process versus another cognitive process when analyzing satellite imagery affects the utility of the output of the analysis. Such tags, when used to tag sufficiently large amouts of process data, can serve as an aid in assessing which types and combinations of cognitive process are more likely to contribute positively or negatively to a decision-making process or mission outcome. Consequently, assessments will also suggest which, and how, cognitive processes should be managed so as to improve the intelligence cycle.

By annotating the investigative processes of analysts using machine readable
ontologies, for example by utilizing logs of their computer operations, we can
collect data both about how intelligence information came to be and also about
what happens to that information in later stages of the intelligence pipeline.
Over time this can be used to improve the intelligence process by using the
ontological annotations as a co-variable to process analysis.
 
The goal of outcomes-based research is to find ways to promote those types of process workflows which have a higher likelihood of generating more useful outcomes. Research of this sort has demonstrated its value most conspicuously in information-driven biomedicine, where relative evaluations of treatment types are generated by associating data about the applications of such treatments to specific patients with data about subsequent outcomes (Clancy and Eisenberg 1998). To implement outcomes-based research in intelligence analysis, we need to make relative evaluations of intelligence processes and process workflows of different types by associating data about instances of such processes with data about subsequent outcomes, and all of this with a focus on cognitive processes and how to stage them into the most effective overall analysis. This requires measuring the intelligence value generated by analytic workflows of given types, using their outcomes to determine average outcomes associated by workflows (cognitive process sequences) of those types, and then drawing conclusions from these average outcomes that allow assignment of metrics. 

\section{An Ontological Strategy}
In building CPO we follow the methodology of ontological realism set forth in (Smith and Ceusters 2010), according to which an ontology should be designed to represent entities in reality, including not only material things, their qualities and functions, and also the information artifacts used to describe and reason about them. As summarized in (CUBRC 2019b):
\begin{quote}
This approach stems from the conviction that disparate ways of capturing data are best rendered interoperable by rendering them conformant to the ways things actually are. Realism implies, then, that any given assertion in an ontology can be evaluated on the basis of an objective criterion: Is the assertion true? … this approach shifts ontology development away from the parochial concerns of particular implementations and toward expanded interoperability.
\end{quote}
Every term in a realist ontology, accordingly, represents a type of entity that is instantiated by real-world instances of this type. The definition of this term captures what is common to all and only instances of this type. This definition has both a natural language form meaningful to human users and a logical form useful for machine processing. By semantically enhancing data with an ontology, both manual and automated methods can be applied to identify the relationships between entities of given types represented by given bodies of data and also to extrapolate new information based on those relationships, thereby complementing the human effort involved in delivering useful intelligence. 

We are aware that many ontology-based approaches to data exploitation have failed. Our approach is based on a methodology developed in the field of bioinformatics, where ontologies -- specifically the Gene Ontology (GO) and a series of ontologies built to interoperate with the Gene Ontology within the so-called OBO (Open Biological Ontologies) Foundry -- are generally recognized as having been successfully applied (Kamdar et al. 2017). This methodology has been piloted for DoD purposes by IARPA and the US Army Research, Development and Engineering Command, which sponsored a process of testing and validating by the International Standards Organization (ISO) and International Electrotechnical Commission (IEC) Joint Technical Committee Metadata Working Group. Two results of this piloting process are of relevance here:
\begin{enumerate}

\item International standard ISO/IEC 21838, approved in 2019 and scheduled to be published in 2020, including ISO/IEC 21838-2: Basic Formal Ontology (BFO), a top-level ontology to promote interoperability of domain ontology development initiatives (ISO/IEC 2020), and
\item The Common Core Ontologies (CCO), a suite of interoperable ontologies based on BFO, including extension ontologies covering many defense and intelligence domains (CUBRC 2019a), are under initial consideration by the Mid-Level Ontology Ad-Hoc Committee of the InterNational Committee for Information Technology Standards (INCITS).

\end{enumerate}

BFO is already being used as top-level architecture for a number of other ontology-building initiatives created under IC auspices, some of which involve use of the CCO, including work by the Applied Physics Laboratory (APL), the Institute for Defense Analysis (IDA) (Chan et al. 2017), and the Science Applications International Corporation (SAIC), as well as by some 300 ontologies development initiatives in medical, scientific and other areas (Smith 2018).

A recent example of the utility of the realist-ontology approach for outcomes-based research  can be seen in (Utecht et al. 2016), which documents the building and implementation of the Ontology of Organizational Structures of Trauma systems and Trauma (OOSTT). By using OOSTT, data is able to be captured in real time and reasoned over by a machine, creating an ever-evolving representation of the domain. The information captured in this process is then able to be exploited to show, for example, correlations between the effectiveness of a trauma center and whether it is following regulations or has a particular organizational structure. Furthermore, the realist approach makes data organized by OOSTT interoperable with other data sets relevant to trauma centers, including data about patient consent (Lin et al. 2014) and patient outcomes (Ceusters and Smith 2006).

\subsection{An Ontology of Internal States}

Almost all DoD- and IC-related efforts in ontology-building, for example (Smith et al. 2013), focus primarily on tagging data collected in external-world areas of interest such as geospatial locations, military units, sensors and their capabilities. Our goal here is to develop a complementary set of ontologies pointing internally, which is to say pointing to the thought processes (and analogous processes inside machines) involved in military and intelligence activities. For our present purposes we focus on the internal processes of single intelligence analysts. More specifically CPO focuses on providing the terminological resources for annotating data about mental processes of this sort which contribute to belief formation and belief changes (for example changes in confidence as to the veridicality of beliefs). 

In building CPO, we reuse terms from existing ontologies wherever possible, but introduce new terms wherever needed. In either case, all terms are defined so as to be compliant with Basic Formal Ontology (BFO) as specified in (Arp et al. 2015). Furthermore, because CPO draws its terms not only from CCO but also from other BFO compliant ontologies, there will be an effort in the interest of interoperability to bring those ontologies, whose content we want to re-use, into the fold of CCO extensions.

\subsection{Analysis as a Feedback Loop}

\begin{figure}
  \includegraphics[width=\linewidth]{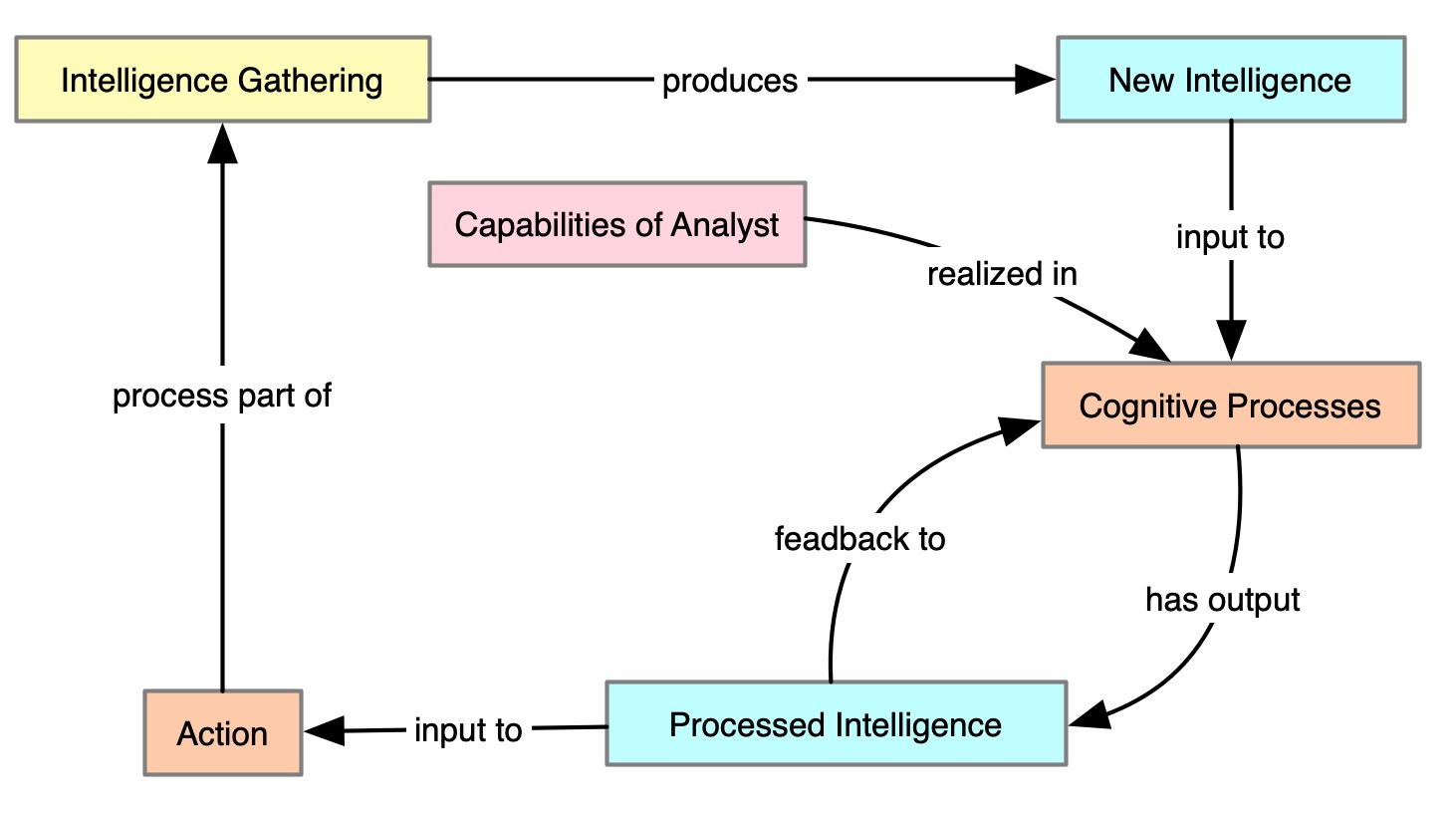}
  \caption{Intelligence Analyst Simple Feedback Loop}
  \label{Fig:Loop}
\end{figure}

The collection and analysis processes we are addressing form feedback loops, as described in (Ford 2010) and as illustrated in Figure \ref{Fig:Loop}. Feedback loops are iterative processes each additional iteration changing based on \textit{feedback} from what happened in previous iterations. The feedback in this case is the intelligence gathered and processed in the prior loop plus any still-relevant intelligence gathered in other previous loops. Gathered intelligence is processed within the loop by cognitive processes -- the realizations of cognitive capabilities -- possessed by analysts and that were largely acquired through training, study, and experience (Merrel et al. 2019). The outcome of processing intelligence includes new beliefs, new hypotheses, changes in confidence -- all of which contribute to a growing body of processed intelligence -- and which precipitate (are inputs to) actions like decisions relating to new intelligence collection steps, queries for further intelligence, or production and dissemination of results.

The proposal is that as the body of processed intelligence grows and various actions are taken, a traceable record of the steps taken within each iteration of a more complex loop will emerge. All paths actually taken through the loop will be recorded and related to outcomes to enable answering of questions like: ``What sorts of queries prove useful in identifying better indicators?'', ``What communication and documentation practices improve the timeliness of analytical output?'', and so on. As the traceable record grows in size it will become available also for more ambitious types of analysis, for example to support the creation of a catalog of lessons learned for use in training future analysts.

Importantly, outcomes include not only the degree to which an analytic workflow led to success or failure but also to what extent, and why, each component of the workflow contributed to this success or failure. For example, queries may be inadequate because they are poorly formed, addressed to the wrong recipient, duplicates, or already issued. Responses to queries, too, may be partial or inadequate. They may, provide only some of the information requested, address indicators incorrectly, provide only partial verification of an indicator, be inconsistent with background assumptions, or be inconsistent with previous responses.

\section{The Cognitive Process Ontology}

The Cognitive Process Ontology (CPO) is inspired by the Mental Functioning Ontology (MF) (Hastings et al. 2012) and builds from the work on representations described in (Smith and Ceusters 2015; Limbaugh et al. 2019; Kasmier et al. 2019).
The central term of CPO -- and the term that generally describes the work of an intelligence analyst -- is `investigative process', a subclass of what the Mental Functioning Ontology terms a `Cognitive Process':
\begin{quote}
Cognitive Process =def. Mental Process that creates, modifies or has as participant some cognitive representation (MF).
\end{quote}
\begin{quote}
Investigative Process =def. Cognitive Process whose agent intends to establish or confirm that some portion of reality exists or does not exist (CPO).
\end{quote}

\noindent An investigative process can be as simple as glancing to confirm the position of the hands of a clock and as complex as an extended International Criminal Police Organization (INTERPOL) terrorist hunt. Importantly, an investigator need not have any idea what she is looking for; she merely needs to be looking for \textit{something}. Investigations unfold as an agent follows \textit{indicators}, which are portions or reality (POR) that affect that agent's estimation that some other portion of reality exists (Bittner and Smith 2008). Practically anything (real) can be a portion of reality, and the same applies to those portions of reality that can serve, theoretically, as indicators. So not only are universals and their instances, including instantiated formal and mathematical relations, portions of reality (and potential indicators), but so are combinations of these, such as a person of interest in Tucson, Italy having a meeting with a known arms dealer at 12pm GMT on October 12, 2014 (Bittner and Smith 2008).
\begin{quote}
  Indicator =def. Portion of Reality that, if it is known to exist, affects our estimation that some other portion of reality exists (CPO). 
\end{quote}

\subsection{Representations}

Mental representations are what we \textit{think with} when performing an investigation. Hence, what we are thinking about is determined by the content of our mental representations. To understand this class let us first introduce `Mental Quality', which is a subclass of BFO `Quality' (ISO/IEC 2020).

\begin{quote}
  Mental Quality =def. Quality which specifically depends on an anatomical structure in the cognitive system of an organism (Smith and Ceusters 2015).
\end{quote}

\noindent We are agnostic as to which parts of an organism constitute its cognitive system. We do however assume that it includes parts of the brain. The term `structure' should also be understood in a very general sense, including for instance areas of the brain with particularly dense neuronal connections specialized to specific mental functioning sorts. 

The definitions of `system' and `cognitive system' presented here are provisional only, and should be read in conjunction with the proposed definition of `bodily system' found in (Smith et al. 2004). 

\begin{quote}
  System =def. Material entity including as parts multiple objects that are causally integrated (Mungall 2019). 
\end{quote}

\begin{quote}
  Cognitive System =def. System which realizes cognitive dispositions, all of whose parts are also parts of a single organism (CPO).
\end{quote}

\noindent As for mental qualities, we remain agnostic as to what their basis might be, that is, what sort of independent continuant they inhere in. Mental qualities are either representational or they are not. Non-representational mental qualities include those that are responsible for giving emotional and sensational processes their characteristic feel. For example, the process of experiencing pain \textit{hurts} because of the mental qualities involved in that process, and similarly for experiences of sorrow or joy. 

\subsubsection{Concretization}

A Representation is a Quality that has information content. Though, in BFO parlance, we would say the Quality \textit{concretizes} information content (ISO/IEC 2020). The significance of concretization is this: if a pattern of qualities concretizes some information content, then producing a similar pattern of qualities spreads that information content to an additional carrier. This is the process commonly referred to as `copying'. Consider the process of copying a quote from a book to a notebook. When copying the quote one reproduces, in the notebook, a patter of qualities similar to that found in the book and, in doing so, spreads the information content from the book to the notebook. When the process completes there are two similar patterns of qualities -- one in the book and one in the notebook -- and the information that was originally carried by the book is now \textit{also} carried by the notebook.

\begin{figure}
  \includegraphics[width=\linewidth]{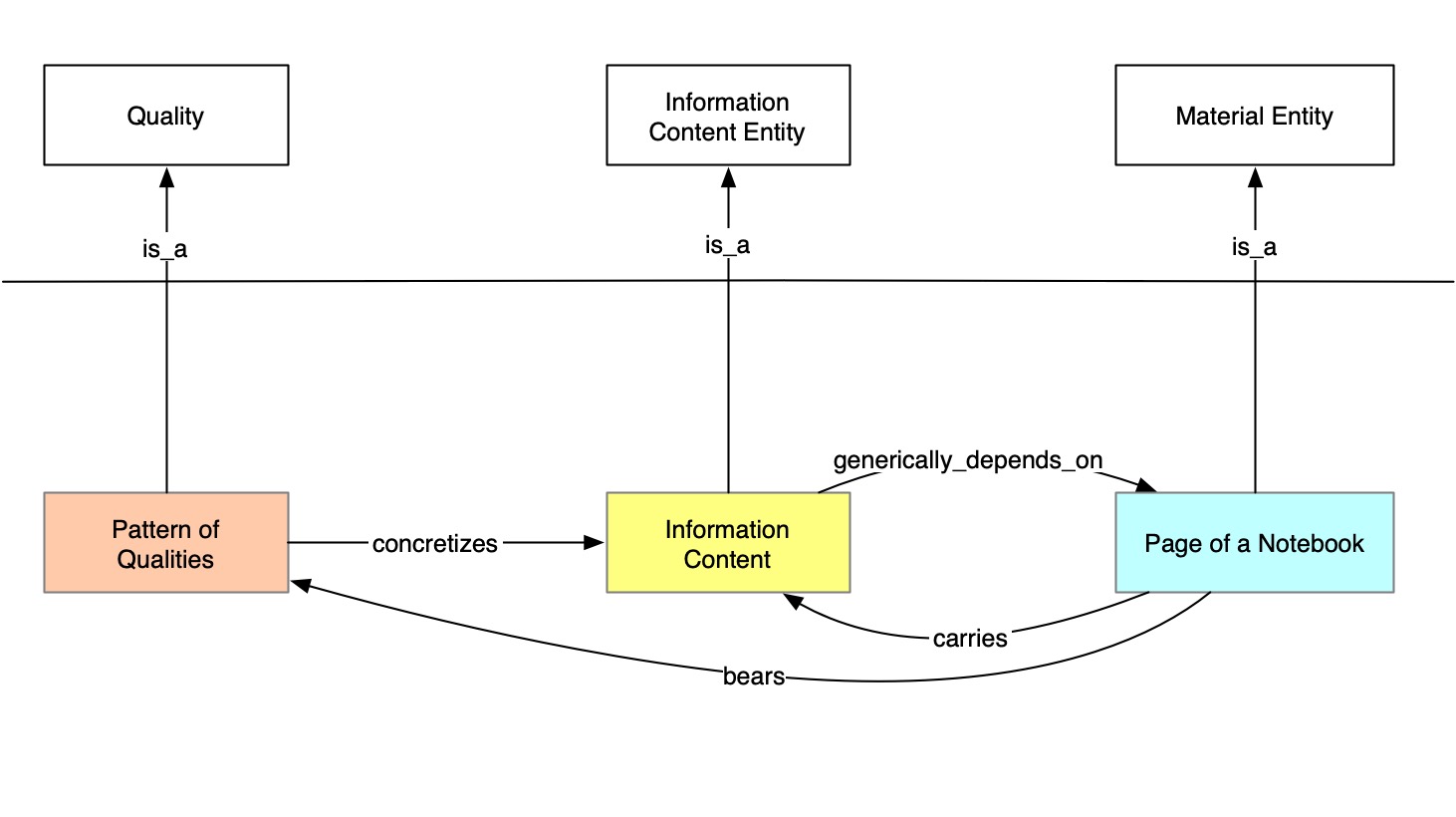}
  \caption{Above the horizontal line are terms from an ontology and below are items of data. A graph that represents a notebook that carries information content.}
  \label{fig:Information}
\end{figure}

According to BFO, items of information content are called `Information Content Entities' (ICEs) and are a subclass of `Generically Dependent Continuant' (GDCs); `generically dependent' here means that an instance of an ICE (or GDC) can have multiple concretizations (ISO/IEC 2020). For example, the particular instance of an ICE that is \textit{Structured Analytic Techniques for Intelligence Analysis} -- also an instance of the subtype \textit{textbook} -- not only exists as concretized by the pattern of qualities inhering in the physical book (made of ink, glue, and paper) on your shelf, but also in the physical book on the shelf in the library, at the used bookstore, as well as concretized by the patterns of electricity that form the pdf file in your laptop. \textit{Structured Analytic Techniques for Intelligence Analysis} is concretized in each case (they are all distinct copies of the same textbook). It is concretized by distinct instances of complex quality patterns inhering in different individual books or digital files. \textit{Structured Analytic Techniques for Intelligence Analysis} thus depends generically on each and every book (or file) that concretizes it, and each and every book (or file) would have to be destroyed to successfully destroy \textit{Structured Analytic Techniques for Intelligence Analysis} itself. In short: an ICE is what various information artifacts can have in common; my copy of \textit{Moby Dick} has in common with your copy of \textit{Moby Dick} its ICE. See figure \ref{fig:Information} for an example of the anatomy of information.

\subsubsection{Mental Representation}\label{5.1.2.}
Mental Representations are Mental Qualities that concretize some ICE. In contrast to MF (Hastings et al. 2012) and earlier versions of CPO (Smith and Ceusters 2015; Limbaugh et al. 2019; Kasmier et al. 2019) we do not here understand Mental Qualities as ever being about anything; this is to increase interoperability with suggested ways of modeling information with CCO (CUBRC 2019c). Rather, the content of a Mental Quality is the ICE that Mental Quality concretizes. 

Mental qualities, in contrast to ICEs, are BFO Specifically Dependent Continuants: features of things that depend for their existence on some bearer. Thus, an instance of a Mental Quality specifically depends on part of a Cognitive System and is only located where its bearer is located, while the ICE concretized by a Mental Quality may also be concretized elsewhere.

When a quality concretizes some ICE, then that quality is a \textit{Representation}, and when that Quality is also of the subtype Mental Quality, then it is a Mental Representation. We define each as follows: 

\begin{quote}
  Representation =def. Quality which concretizes some Information Content Entity (CPO).
\end{quote}

\begin{quote}
  Mental Representation =def. Representation which is a Mental Quality (Smith and Ceusters 2015).
\end{quote}

\noindent Mental Qualities are never strictly speaking about anything. However, Mental \textit{Representations}, because of the content they concretize, are responsible for the intentionality (or directedness) found in a Cognitive Process. When asked, ``What are you thinking about?'' the answer is dependent on the Mental Representations had by the parts of the cognitive system that are participating in that process. 

\subsubsection{Two Types of Concretization}

We distinguish two types of concretization: \textit{original} and \textit{derived} (compare Searle on `intrinsic' and `derived' intentionality in (Searle 1983)). This distinction mirrors that between bona fide and fiat boundaries; both types of boundaries exist and are genuine, but the former are associated with physical discontinuities, like walls and rivers, while the latter come into existence only through the intentional actions of agents, such as the signing of a legal document that specifies property lines (Smith 2001). Though the resultant property lines are products of fiat, they are nonetheless parts of reality and have legal significance. 

\subsubsection{Original Concretizers}
Entities that are the original concretizers of some ICE are various types of Mental Representations. The concretization here is \textit{original} because, like bona fide boundaries, the concretization relation between a mental representation and its ICE is in no way derived from the intentions of agents; that is, it is in no way derived from the way agents repeatedly use some symbol to communicate content; it is always the other way around.

\subsubsection{Derived Concretizers}

The paradigm entities that concretize derivatively are non-mental representations, as in symbols (quality-patterns) such as `dog' or `$\pi$', either spoken, written, or otherwise instantiated outside of the mind (if there truly are meaningless \textit{symbols}, then they would not be concretizers). A symbol begins to concretize some ICE because of how it is used (or intended to be used) by some intentional agent. Thus, it was only after an act of naming that the symbol `$\pi$' became one way of expressing the ICE otherwise expressed as the symbol `pi'.

Following Chisholm's doctrine of the primacy of the mental (Chisholm 1984), the existence of a non-mental representation, which always concretizes its ICE derivatively, is explained in terms of the original concretization of that ICE by some Mental Representation and the intended use of symbols, by some agent, to increase the number of carriers of that ICE. (In what follows, if an expression is surrounded by single quotes (`some expression'), then it denotes a \textit{symbol}, and if surrounded by double quotes (``some expression''), then the \textit{content} of some set of symbols is denoted instead.) The reason `$\pi$' is associated with the ICE ``the ratio of the circumference and the diameter of a circle'' is because of, first, the original concretization of ``the ratio of the circumference and the diameter of a circle'' in the mind of William Jones, who first introduced the symbol to carry the same ICE as `pi', and later the original concretization in the minds of nearly every student who learned the language of mathematics. Before the intention of Jones for `$\pi$' to be a vehicle for an ICE, `$\pi$' was merely part of the Greek alphabet. Though original concretization always explains derived concretization, the temporal order can sometimes be backwards-looking. A photograph taken by a motion-sensor-activated camera concretizes an ICE about \textit{the intruder}, not because we baptized it as such, but because we recognize it as such.

\subsubsection{Cognitive Representation}

Cognitive Representation is a subtype of Mental Representation (defined in \ref{5.1.2.} above) instances of which always concretize some Descriptive ICE. As the concretizers of descriptions, Cognitive Representations (more specifically their content) have what Searle called a `mind-to-world direction of fit' (Searle 2001; Anscombe 1963). That is, the accuracy of a Cognitive Representation is measured according to how well its descriptive content matches reality. If a Cognitive Representation is inaccurate, then the error is in the Cognitive Representation and not elsewhere: a Cognitive Representation aims to fit what it is intended to be about \textit{in the world} and not vice versa. (While here we speak of a Cognitive Representation's being correct, its degree of accuracy, and what it is about, each of these is derivative on the representation's concretized ICE. A strategy for preserving interoperability between graphs that make the derivative nature of this relationship explicit and those that do not can be found here (Dodds and Davis 2020).) 

A Cognitive Representation, then, is entirely accurate (henceforth; `veridical') when it is about \textit{the} portion of reality that its content is intended to be about (Smith and Ceusters 2015). If I think of the mug on my desk as green, but upon entering my office discover that it is red, then my Cognitive Representation of the mug is not veridical (thought it may still be accurate to some extent in virtue of there actually being a mug on my desk and not, say, a carafe). Thus, a Cognitive Representation is veridical when the represented POR -- the POR the content of the Cognitive Representation is about -- exists as it is represented. 

The phrase `as represented' implies that there are different granularities of representation. That is, Cognitive Representations can be more or less detailed. For example, were Cognitive Representation\_1 of \textit{the mug's being green}, but not of the mug's being any \textit{specific} shade of green, then the mug's being merely green, regardless of the shade, would be enough to make that Cognitive Representation\_1 veridical.

\begin{quote}
  Cognitive Representation =def. Mental Representation that has a mind-to-world direction of fit (CPO).
\end{quote}

\noindent Contrast this with a type of Mental Representation that would be associated with a desire; a desire demands (so to speak) that the world fit it and not vice versa; it has a world-to-mind direction of fit. For example, say I want the mug on my desk to be green even though it is red. The world is wrong according to my \textit{want}, which is to say that the word is wrong according to to a Mental Representation which has a world-to-mind direction of fit.

\subsubsection{Representation that is Believed}

Some Cognitive Representations are taken by the agent to be veridical, whether they are actually veridical or not. Such a Cognitive Representation is what we referred to above with the term `Representation that is Believed' (RTB). An RTB is treated by the agent (by her Cognitive System) as actually true, though it may or may not be actually true. More specifically, what distinguishes an RTB from a mere Cognitive Representation is that the latter is fused with a Positive Confidence Value. (Compare what Meinong has to say about Ernstgefühle or serious (or earnest) mental phenomena in (Meinong 1907).) 

`Fusion' is a term adapted from Husserl (Husserl 1970) and is a primitive relationship that obtains between multiple Quality instances when they are so closely related that an additional Quality instance seems to emerge from them. Take for example what appears to be a solid green image displayed on a television screen, which upon very close inspection is actually colored by means of tiny yellow and blue squares, or pixels, thus giving a green appearance to the naked eye. The pixels are bearers of many instances of yellow and blue, and these instances appear to have fused into an additional instance of greenness. Similarly, when an instance of a Cognitive Representation and an instance of Positive Confidence Value are fused together in a cognitive system there seems to be an additional quality instance: an instance of an RTB. (In the parlance of the Web Ontology Language (OWL), the `is fused with' relation is way to build \textit{defined classes} out of Specifically Dependent Continuants.)

A \textit{Confidence Value} is a non-representational mental quality that, when fused with a Cognitive Representation, determines how that Cognitive Representation is utilized by a Cognitive System. If a Confidence Value is measured as `positive', then its fused Cognitive Representation is treated as more-likely-than-not veridical, and if the Confidence Value is negative, then the Cognitive Representation is treated as more-likely-than-not not-veridical. 

For example, if Cognitive Representation CR2 that `My coffee is still too hot to drink' is fused with a positive Confidence Value, then were CR2 taken as input by the agent's Cognitive System when deciding to take a sip of the coffee, then because of CR2's influence, the agent would blow on her coffee first before taking a sip. Importantly, a fused Confidence Value should not be confused with second-order Cognitive Representations, such as a Representation about the likelihood of another Representation's being veridical (as for example when you are asked: ``Are you sure?'') Such second-order Mental Representations are distinct from the pre-introspective and non-representational confidence that we find fused with those Cognitive Representations which are RTBs. 
\begin{quote}
  Confidence Value =def. Mental Quality that, when fused with a Cognitive Representation CR, determines the extent to which a Cognitive System operates as if CR is veridical (CPO).
\end{quote}

\noindent With this in mind we can now define `Representation that is Believed' as follows: 

\begin{quote}
  Representation that is Believed (RTB) =def. Cognitive Representation that is fused with a positive Confidence Value (CPO).
\end{quote}

It might be objected that a positive Confidence Value is not enough to render a representation \textit{believed.} For example, if the Confidence Value is just barely above the more-likely-than-not threshold, then the representation is too uncertain to be rightly said to be \textit{believed}. This objection confuses the project of applied ontology, which is to use terms to carve out relevant portions of reality, with linguistics. `Representation that is Believed' is not meant to be a synonym of the term `belief' as used in everyday discourse. This is in part because there is likely no sufficiently consistent natural language specification of the meaning of the term `belief' that is also useful. Importantly, there is also likely no sufficiently consistent natural language use of terms, such as `Disposition', `Function', `Disease', `Disorder' and many other basic ontological terms. Like `Representation that is Believed' these terms do not attempt to be synonyms with their natural language counterparts, and are instead seen as carving out relevant portions of reality that are akin to those referred to by their natural language counterparts.

\subsubsection{Representation that is Warranted}
Following Plantinga (Plantinga 1993), a Representation that is Warranted (RTW) is an RTB which holds an epistemically privileged place in a Cognitive System: what in other contexts we might call an `item of knowledge'. It is so privileged because it was produced by some designed or vetted process so that, when in an environment of the sort that it was designed or vetted for, it reliably outputs veridical Cognitive Representations. As an analogy, consider an algorithm whose functioning is only designed and vetted for reliability for a certain type of input data. That algorithm would be functioning properly only when processing data of that type. Furthermore, the outputs of that algorithm should only be trusted when they are the product of the algorithm when it is functioning properly (or, of course, when the outputs are independently verified). As such, if an RTB is produced through a process of proper cognitive functioning, then it isn't just \textit{de facto} fused with a positive confidence value but also is such that it \textit{should} be fused with a positive confidence.

Instances of such processes are instances of `Process of Proper Cognitive Functioning' (PPCF), and the RTB's formed by such processes are \textit{warranted}:
\begin{quote}
  Process of Proper Cognitive Functioning (PPCF) =def. Cognitive Process that has been successfully vetted or designed to reliably form veridical Cognitive Representations in environments of given types that include the environment in which the Cognitive Process is occurring (CPO).

\end{quote}
\begin{quote}
  Representation that is Warranted (RTW) =def. Representation that is Believed formed through Proper Cognitive Functioning in its vetted- or designed-for environment (CPO).

\end{quote}

\subsubsection{Representing Features of Cognitive Representations}

\begin{figure}
  \includegraphics[width=\linewidth]{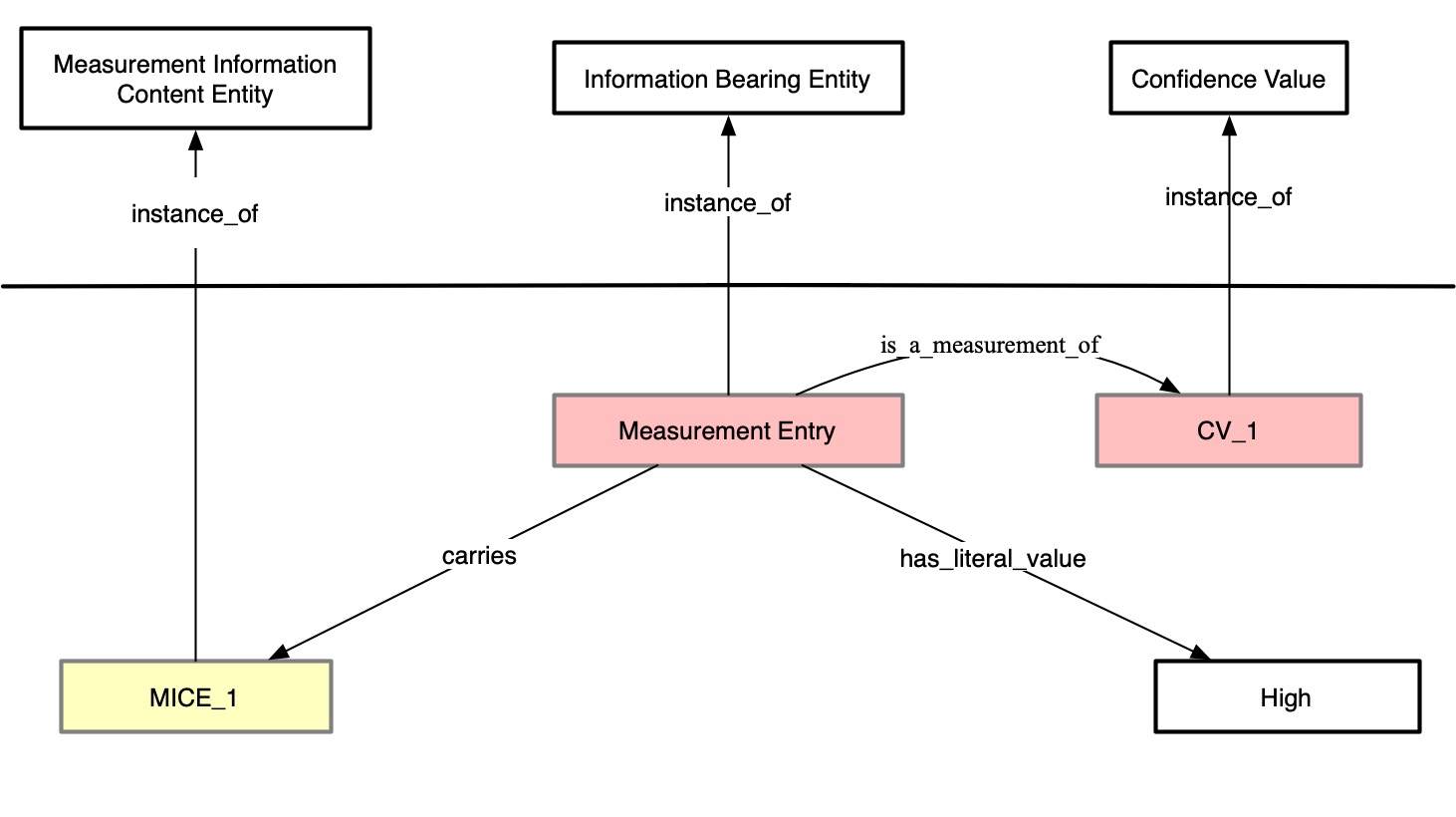}
  \caption{Above the horizontal line are terms from an ontology and below are items of data. A graph schema of a measurement.}
  \label{fig:Measurement}
\end{figure}

Mental phenomenon like the accuracy of a Cognitive Representation or the strength of a Confidence Value are normative phenomenon; that is, they can be graded as better or worse according to a standard. We represent the grade of such phenomenon as a measurement. Following CCO, and as shown in Figure \ref{fig:Measurement} the graph representation of a measurement has three nodes, 
\begin{enumerate}

\item a \textit{Measurement Information Content Entity} (MICE), 
\item the \textit{carrier} of that Measurement Information Content Entity, called an `Information Bearing Entity', which is always a Material Continuant, and
\item the \textit{literal value} of the carrier (that which the content generically depend on), which tells us what the carrier would look like, and thus how it would be read. \end{enumerate}

\noindent Definitions are as follows:

\begin{quote}
 Measurement Information Content Entity (MICE) =def. Descriptive Information Content Entity that describes the extent, dimensions, quantity, or quality of an Entity relative to some standard (CCO).
\end{quote}

\begin{quote}
Information Bearing Entity (IBE) =def. Object upon which an Information Content Entity generically depends (CCO).
\end{quote}

\noindent Furthermore, literal values are related to information carriers by OWL data properties like, `has boolean value', `has nominal value', and `has decimal value'.

\begin{figure}
  \includegraphics[width=\linewidth]{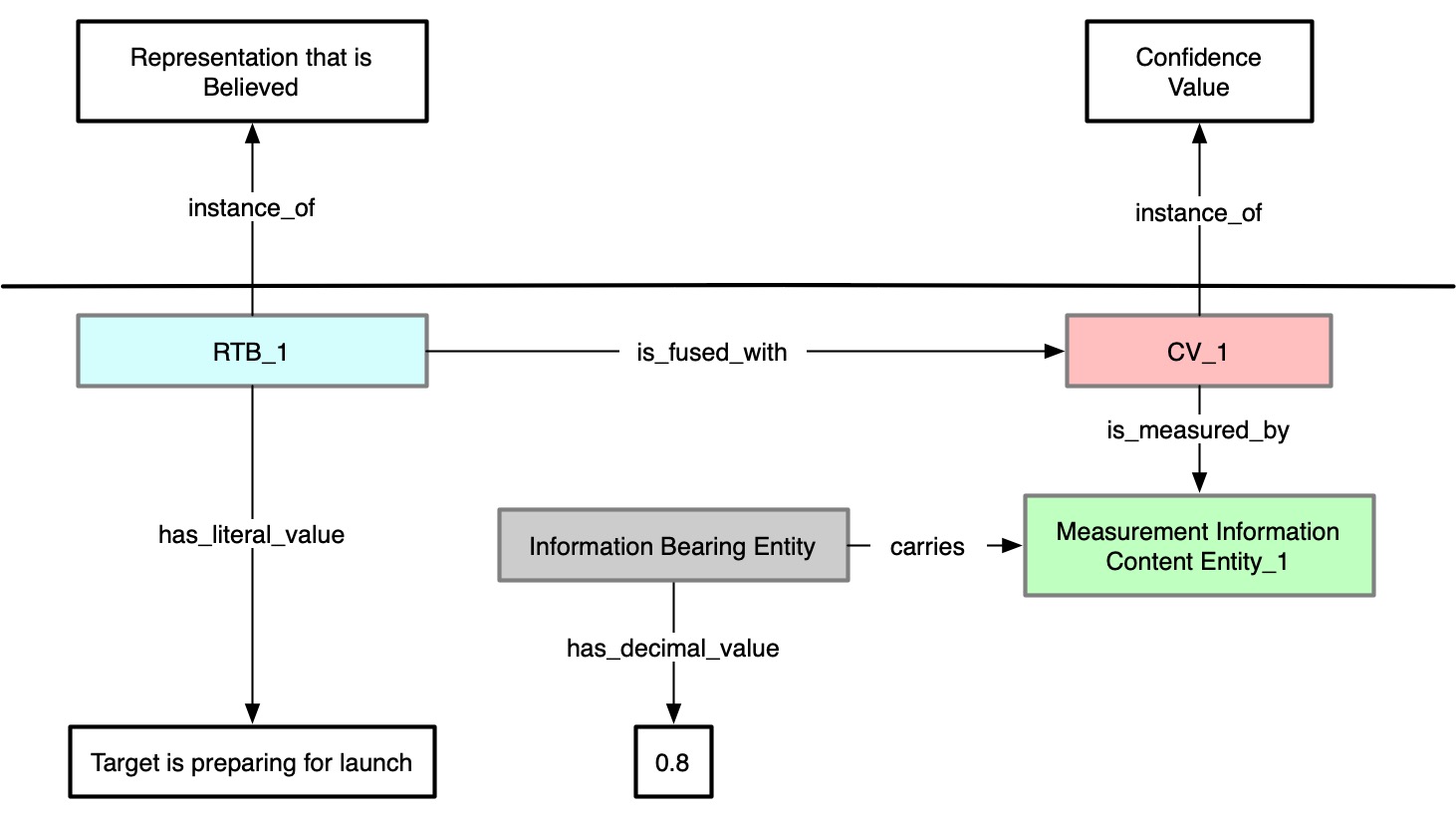}
  \caption{Above the horizontal line are terms from an ontology and below are items of data. A graph schema of a Representation that is Believed.}
  \label{fig:RTB}
\end{figure}

Using this schema, we represent Representations that are Believed as Cognitive Representations that are fused with a Confidence Value measured as positive; in Figure \ref{fig:RTB} we use a grade of `0.8' to indicate positive confidence. Furthermore, we represent the literal value of a Cognitive Representation itself as if the content of that representation were concretized outside of the mind; for instance, were it written down in a sentence or in the form of a photograph. (This is an imperfect solution demanded by private nature of mental phenomenon.)

\begin{figure}
  \includegraphics[width=\linewidth]{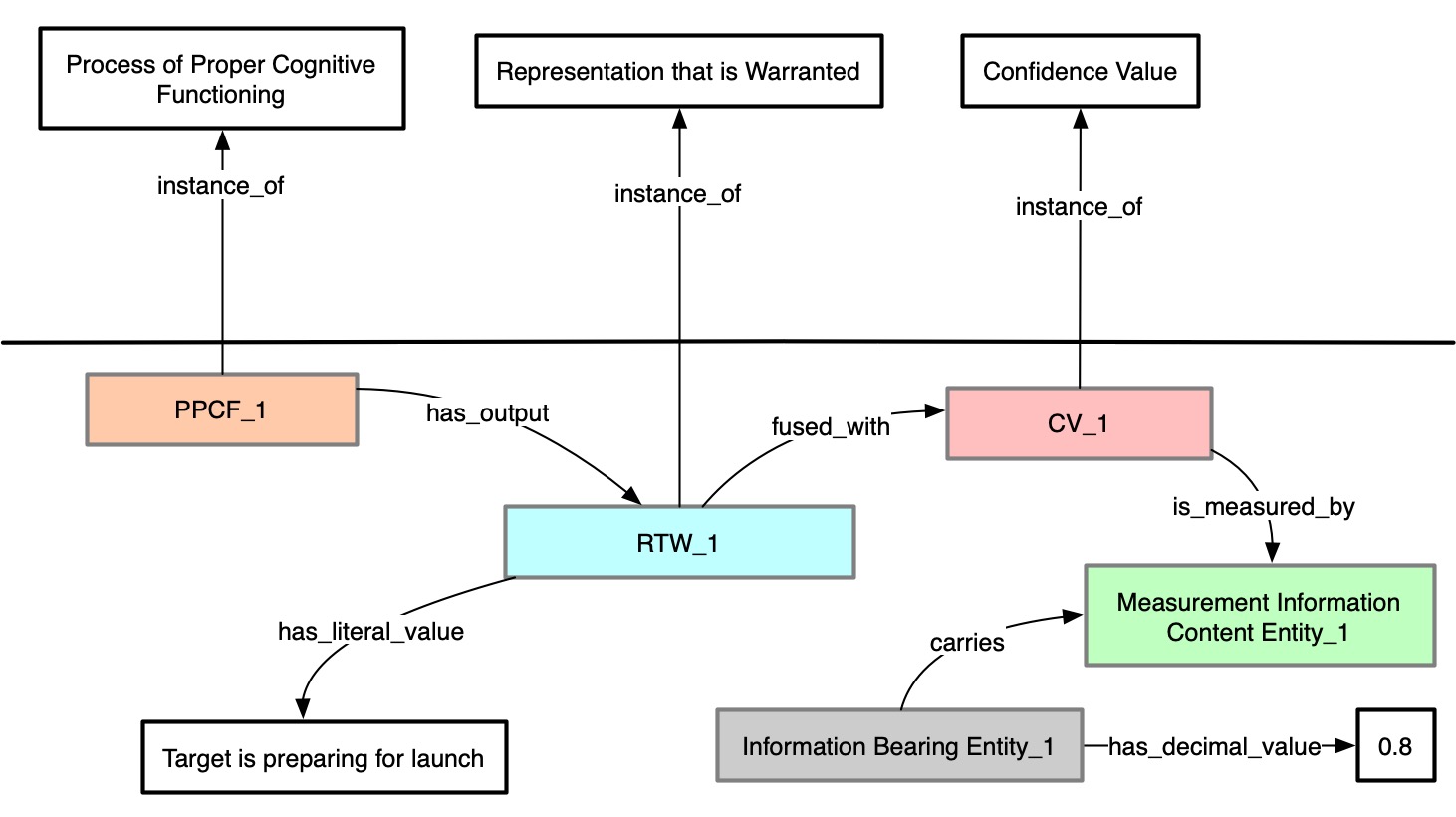}
  \caption{Above the horizontal line are terms from an ontology and below are items of data. A graph of a Representation that is Warranted including its entailments.}
  \label{fig:RTW}
\end{figure}

We represent RTWs as RTBs that are the output of PPCFs, as in Figure \ref{fig:RTW}. The privilege of an RTW is that it can justifiably be used in a Cognitive Process without further scrutiny. This is because an instance of an RTW, by definition, is 1) produced by some PPCF and thus 2) rightly fused with a positive Confidence Value, which in this case is indicated by a decimal value representing a purported high chance (80\%) of veridicality. Because these components are definitional, they can be enforced by reasoners, and thus it can be regulated that something only be tagged as an RTW if it has the requisite relations. 

\begin{figure}
  \includegraphics[width=\linewidth]{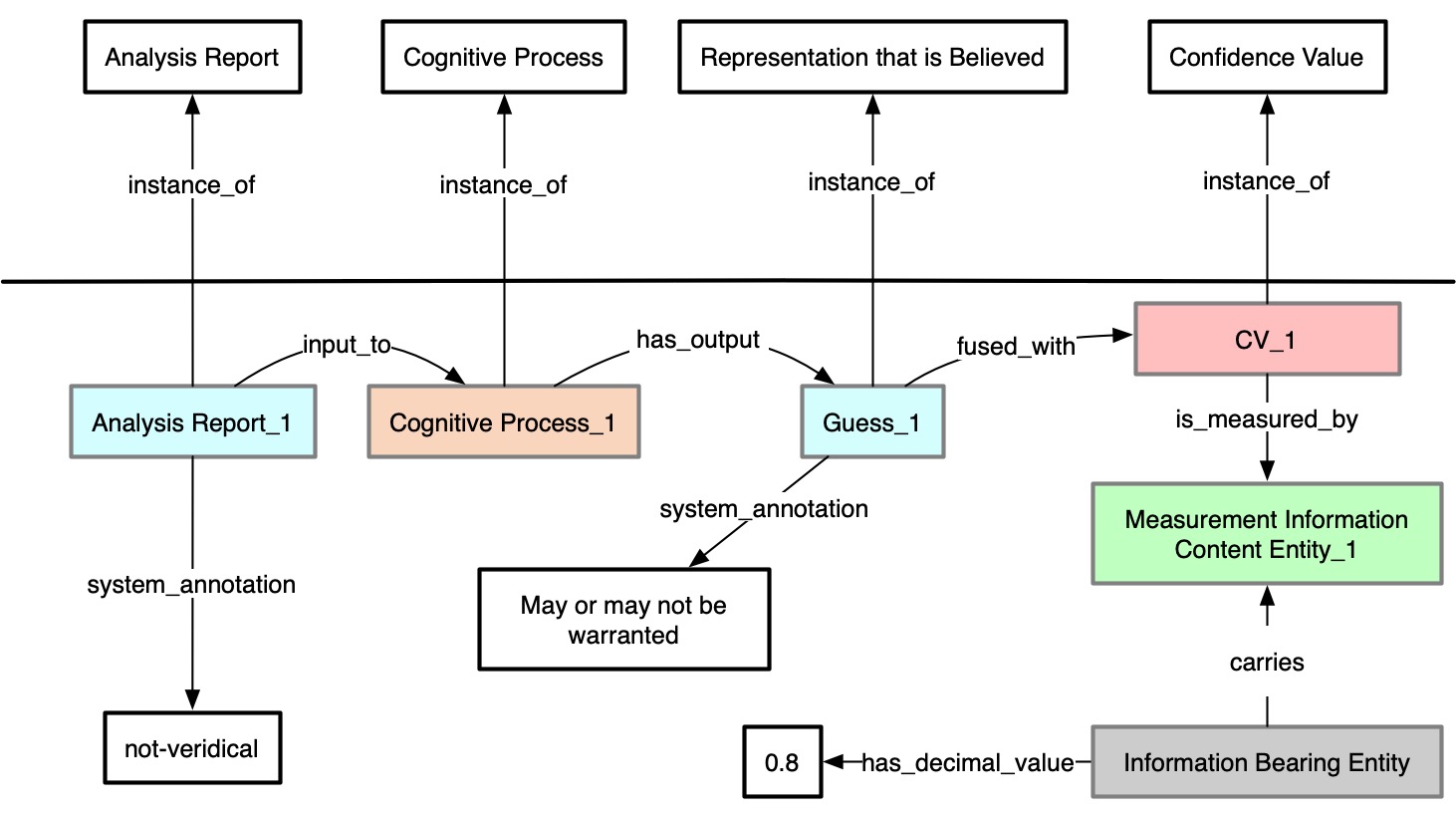}
  \caption{Above the horizontal line are terms from an ontology and below are items of data. A graph of a Representation that is Believed that is a mere guess because it was not formed through proper cognitive functioning due to the source information being not-veridical.}
  \label{fig:Warrant_Demo}
\end{figure}

Furthermore, the introduction of warrant allows for a dimension of data integrity that goes beyond veridicality or confidence. For example, Figure \ref{fig:Warrant_Demo} illustrates the formation of an RTB that may or may not be veridical but which the user holds with high confidence, although the Cognitive Process that outputted the RTB is not a PPCF. Here it is assumed that the relevant type of Cognitive Process requires veridical input data to be reliable; as such, even though the analyst may represent her RTB as fused with a high confidence value, the system knows to explicitly represent the information -- through a system annotations -- as unwarranted: a mere guess (compare with (Hogan and Ceusters 2016)). The upshot is that warrant, veridicality, and confidence can each provide a dimension of data integrity to use when assessing information for the sake of decision making and outcomes-based research. 

\section{Using CPO}

How can CPO be used in practice? To CPO-tag the analysis process, all activities performed by intelligence analysts or systems, would be tagged with CPO terms. For system activities, which are always deterministic, this can be achieved by using mapping tables indicating which activity to tag with which terms. For human user activities, such tables can be used whenever the user performs activities for which enumerations or numerical values can natively be provided to the system. These are activities such as clicking buttons or selecting choices from drop-down menus or entering numerical parameters.

For free-text entries into the system, such tables would not work. Instead, the user could either be asked to select CPO-tagging terms from a menu (which does not work well in practice, as evidenced for example from similar mechanisms in hospital electronic health record systems). A better alternative, however, would be to tag texts using stochastic or deterministic natural language processing (NLP).

Once such tags have been made available, the process flow in each unit could be analyzed. Process-evaluation via regressions between the process steps and its outcomes could be performed to identify opportunities for continuous improvement. Furthermore, process flows revealing analysis style and efficacy could be compared between units using multivariate temporal process pattern analysis techniques to establish gold standards and contribute to the discovery and dispersion of the best approaches. The result is advanced sabermetrics but for the IC and DOD communities (Lewis 2003).

Using CPO to train machine learning algorithms can be excluded for two reasons. First of all, the cognitive tasks performed by humans in intelligence data analysis cannot be modeled using mathematical models (Landgrebe and Smith 2019b). And even if they could be so modeled, the granularity of the data points provided by CPO-tags would never suffice to obtain the necessary input matrix density: most of the process steps in intelligence analysis are implicit. Nevertheless, an automated tagging with CPO could contribute to an increase in the effectiveness and the efficacy of intelligence processes.

\section{Conclusion}

We have proposed a research program for using CPO to enhance the scalability and interoperability of analyst data. However, we have also laid out a practical and ontologically sound theory of a type of knowledge that we call `Representation that is Warranted'. The significance of this is that it carves a path to assess the integrity of items of data along a dimension other than veridicality. Future work will see the application of `warrant' beyond mental phenomenon allowing us to ask also of a document such as an intelligence report, not only whether it is veridical, but also whether it should (or should have been) trusted. Once this has been achieved we will then be in a position to explore the integration of this dimension of integrity into our investigative pipelines.\\

\noindent \textit{This research was supported by an appointment held by David Limbaugh to the Intelligence Community Postdoctoral Research Fellowship Program at the University at Buffalo, administered by Oak Ridge Institute for Science and Education through an interagency agreement between the U.S. Department of Energy and the Office of the Director of National Intelligence.}

\end{document}